\documentclass[11pt]{article}
\usepackage{amsmath}
\usepackage{acl2015}
\usepackage{tabularx}
\usepackage{times}
\usepackage{url}
\usepackage{latexsym}
\usepackage{amsfonts}
\usepackage{mathtools}
\usepackage{tikz}
\usepackage{adjustbox}
\usepackage{algorithm}
\usepackage{algorithmic}
\usetikzlibrary{shapes, arrows, positioning}
\definecolor{forestgreen}{rgb}{0.13,0.54,0.13}

\newcommand{\cev}[1]{\reflectbox{\ensuremath{\vec{\reflectbox{\ensuremath{#1}}}}}}
\newcommand{\wtvTitle}[2]{{{\operatorname{#1}}_{\operatorname{#2}}}}
\newcommand{\wtv}[2]{ \vec{\vphantom{l}\operatorname{#1}}_{\operatorname{#2}} }
\newcommand{\wtc}[2]{ \cev{\vphantom{l}\operatorname{#1}}_{\operatorname{#2}} }

\title{Trans-gram, 
    Fast Cross-lingual Word-embeddings \\
    {$\wtvTitle{rey}{es} - \wtvTitle{Mann}{de} = \wtvTitle{regina}{it} - \wtvTitle{femme}{fr} $}
}

\author{
  Jocelyn Coulmance \\
  105 rue La Fayette \\
  75010 Paris \\
  {\tt joc@proxem.com}
  \And
  Jean-Marc Marty \thanks{\: These authors contributed equally.} \\
  105 rue La Fayette \\
  75010 Paris \\
  {\tt jmm@proxem.com}
  \And
  Guillaume Wenzek \footnotemark[1]\\
  105 rue La Fayette \\
  75010 Paris \\
  {\tt guw@proxem.com}
  \And
  Amine Benhalloum \\
  105 rue La Fayette \\
  75010 Paris \\
  {\tt aba@proxem.com}
}
\date{}

\begin{document}
\maketitle
\begin{abstract}
We introduce \emph{Trans-gram}, a simple
and computationally-efficient method to simultaneously learn and
align word-embeddings for a variety of languages, using only
monolingual data and a smaller set of sentence-aligned data.
We use our new method to compute aligned word-embeddings for twenty-one languages using English as a pivot language. We show that some linguistic features are aligned across languages for which we do not have aligned data, even though those properties do not exist in the pivot language.
We also achieve state of the art results on standard cross-lingual text classification and word translation tasks.
\end{abstract}

\section{Introduction}
Word-embeddings are a representation of words with fixed-sized vectors.
It is a distributed representation \cite{hinton1984distributed} in the
sense that there is not necessarily a one-to-one correspondence
between vector dimensions and linguistic properties.
The linguistic properties are distributed along the dimensions of the space.

A popular method to compute word-embeddings is the Skip-gram model \cite{mikolov2013efficient}. This algorithm learns high-quality word vectors with a computation cost much lower than previous methods. 
This allows the processing of very important amounts of data. For instance, a 1.6 billion words dataset can be processed in less than one day.

Several authors came up with different methods to align word-embeddings across two languages \cite{klementiev2012inducing,mikolov2013exploiting,lauly2014autoencoder,gouws2014bilbowa}. 

In this article, we introduce a new method called Trans-gram, which learns word embeddings aligned across many languages, in a simple and efficient fashion, using only sentence alignments rather than word alignments.
We compare our method with previous approaches on a cross-lingual document classification task and on a word translation task and obtain state of the art results on these tasks.
Additionally, word-embeddings for twenty-one languages are learned simultaneously - to our knowledge - for the first time, in less than two and a half hours.
Furthermore, we illustrate some interesting properties that are captured such as cross-lingual analogies, e.g 
$\wtv{rey}{es} - \wtv{Mann}{de} + \wtv{femme}{fr} \approx \wtv{regina}{it}$
which can be used for disambiguation.

\section{Review of Previous Work}

A number of methods have been explored to train and align bilingual word-embeddings.
These methods pursue two objectives: 
first, similar representations (i.e. spatially close) must be assigned to similar words (i.e. ``semantically close'') within each language - this is the \textbf{mono-lingual objective};
second, similar representations must be assigned to similar words across languages - this is the \textbf{cross-lingual objective}. 

The simplest approach consists in separating the mono-lingual optimization task from the cross-lingual optimization task.
This is for example the case in  \cite{mikolov2013exploiting}.
The idea is to separately train two sets of word-embeddings for each language and then to do a parametric estimation of the mapping between word-embeddings across languages.
This method was further extended by \cite{faruqui2014improving}.
Even though those algorithms proved to be viable and fast, it is not clear whether or not a simple mapping between whole languages exists.
Moreover, they require word alignments which are a rare and expensive resource. 

Another approach consists in focusing entirely on the cross-lingual objective.
This was explored in \cite{hermann2013multilingual,lauly2014autoencoder} where every couple of aligned sentences is transformed into two fixed-size vectors.
Then, the model minimizes the Euclidean distance between both vectors.
This idea allows processing corpus aligned at sentence-level rather than word-level.
However, it does not leverage the abundance of existing mono-lingual corpora .

A popular approach is to jointly optimize the mono-lingual and cross-lingual objectives simultaneously.
This is mostly done by minimizing the sum of mono-lingual loss functions for each language and the cross-lingual loss function.
\cite{klementiev2012inducing} proved this approach to be useful by obtaining state-of-the-art results on several tasks.
\cite{gouws2014bilbowa} extends their work with a more computationally-efficient implementation.

\section{From Skip-Gram to Trans-Gram}

\subsection{Skip-gram}

We briefly introduce the Skip-gram algorithm, as we will need it for further explanations. Skip-gram allows to train word embeddings for a language using mono-lingual data.
This method uses a dual representation for words.
Each word $w$ has two embeddings: a target vector, $\vec{w}$ ($\in \mathbb{R}^{D}$), and a context vector, $\cev{w}$ ($\in \mathbb{R}^{D}$).
The algorithm tries to estimate the probability of a word $w$ to appear in the context of a word $c$.
More precisely we are learning the embeddings $\vec{w}$, $\cev{c}$ so that:
$\sigma(\vec{w} \cdot \cev{c}) = P(w|c)$ where $\sigma$ is the sigmoid function.

A simplified version of the loss function minimized by Skip-gram is the following:
\begin{equation}
    J = \displaystyle 
        \sum_{s \in C}{
            \sum_{w \in s}{
                \sum_{c \in s[w-l:w+l]}{
                    - \log \sigma (\vec{w} \cdot \cev{c})
                }
            }
        }
\end{equation}
where $C$ is the set of sentences constituting the training corpus, and $s[w-l:w+l]$ is a word window on the sentence $s$ centered around $w$.
For the sake of simplicity this equation does not include the ``negative-sampling'' term, see \cite{mikolov2013efficient} for more details.



Skip-gram can be seen as a materialization of the distributional hypothesis \cite{harris1968mathematical}: ``Words used in similar contexts have similar meanings''.
We will now see how to extend this idea to cross-lingual contexts.


\subsection{Trans-gram}
In this section we introduce Trans-gram, a new method to compute aligned word-embeddings for a variety of languages.

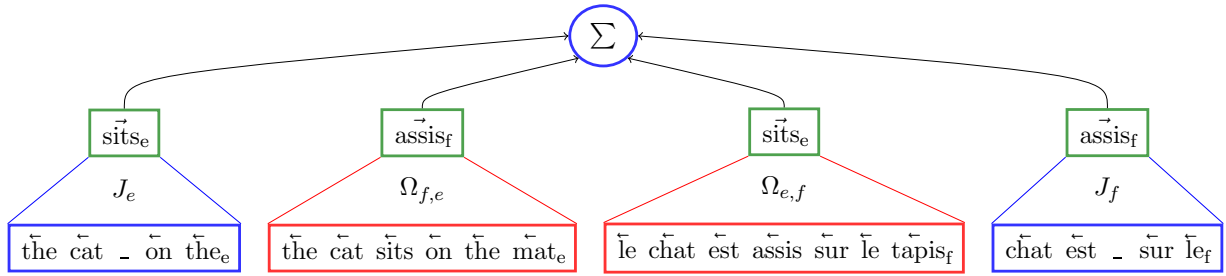
\begin{figure*}[t]
\begin{adjustbox}{max width=\textwidth}
\begin{tikzpicture}
[
sum/.style={ellipse, draw=blue!80, fill=white!0, very thick, minimum width=1cm},
wordSkip/.style={rectangle, draw=forestgreen!80, fill=white!0, very thick, minimum width=1cm},
sentSkip/.style={rectangle, draw=blue!80, fill=white!0, very thick,  minimum width=1cm, xshift=-6mm},
arrowSkip/.style={draw=blue},
wordTrans/.style={rectangle, draw=forestgreen!80, fill=white!0, very thick, minimum width=1cm},
sentTrans/.style={draw=red!80, fill=white!0, very thick,  minimum width=1cm, xshift=-6mm},
arrowTrans/.style={draw=red}
]

\node[sentSkip]  (sEn)    {$\wtc{the}{}$ $\wtc{cat}{}$ \;\textunderscore\; $\wtc{on}{}$ $\wtc{the}{e}$};
\node[wordSkip]  (satEn)  [above = of sEn]    {$\wtv{sits}{e}$};
\draw[arrowSkip] (sEn.north west)    --  (satEn.south west);
\draw[arrowSkip] (sEn.north east)    --  (satEn.south east);
\path (sEn.north) -- node (JEn) {$J_{e}$} (satEn.south);

\node[sentTrans]  (sFrEn)   [right = of sEn]   {$\wtc{the}{}$ $\wtc{cat}{}$ $\wtc{sits}{}$ $\wtc{on}{}$ $\wtc{the}{}$ $\wtc{mat}{e}$};
\node[wordTrans]  (satFrEn) [above = of sFrEn]      {$\wtv{assis}{f}$};
\draw[arrowTrans] (sFrEn.north west)    --  (satFrEn.south west);
\draw[arrowTrans] (sFrEn.north east)    --  (satFrEn.south east);
\path (sFrEn.north) -- node (JFrEn) {$\Omega_{f,e}$} (satFrEn.south);

\node[sentTrans]  (sEnFr)   [right = of sFrEn]   {$\wtc{le}{}$ $\wtc{chat}{}$ $\wtc{est}{}$ $\wtc{assis}{}$ $\wtc{sur}{}$ $\wtc{le}{}$ $\wtc{tapis}{f}$};
\node[wordTrans]  (satEnFr) [above = of sEnFr]   {$\wtv{sits}{e}$};
\draw[arrowTrans] (sEnFr.north west)    --  (satEnFr.south west);
\draw[arrowTrans] (sEnFr.north east)    --  (satEnFr.south east);
\path (sEnFr.north) -- node (JEnFr) {$\Omega_{e,f}$} (satEnFr.south);

\node[sentSkip]  (sFr)   [right = of sEnFr] {$\wtc{chat}{}$ $\wtc{est}{}$ \;\textunderscore\; $\wtc{sur}{}$ $\wtc{le}{f}$};
\node[wordSkip]  (satFr) [above = of sFr]   {$\wtv{assis}{f}$};
\draw[arrowSkip] (sFr.north west) --  (satFr.south west);
\draw[arrowSkip] (sFr.north east) --  (satFr.south east);
\path (sFr.north) -- node (JFr) {$J_f$} (satFr.south);

\path (satFrEn.base) -- node (b) {} (satEnFr.base);
\node[sum]  (sum) [above = of b]   {$\sum$};
\draw[->] (satEn.north) .. controls +(up:5mm) ..  (sum.west);
\draw[->] (satFrEn.north) .. controls +(up:2mm) ..  (sum.south west);
\draw[->] (satEnFr.north) .. controls +(up:2mm) ..  (sum.south east);
\draw[->] (satFr.north) .. controls +(up:5mm) ..  (sum.east);

\end{tikzpicture}
\end{adjustbox}

\caption{
\fontsize{8}{4} The four partial objectives contributing to the alignment of English and French: a Skip-gram objective per language ($J_e$ and $J_f$) over a window surrounding a target word (blue) and two Trans-gram objectives ($\Omega_{e,f}$ and $\Omega_{f,e}$) over the whole sentence aligned with the sentence from which the target word is extracted (red).
}
\label{trans_gram}
\end{figure*}



Our method will minimize the summation of mono-lingual losses and cross-lingual losses. Like in BilBOWA \cite{gouws2014bilbowa}, we use Skip-gram as a mono-lingual loss.
Assuming we are trying to learn aligned word vectors for languages $e$ (e.g. English) and $f$ (e.g. French), we note $J_e$ and $J_f$ the two mono-lingual losses.

In BilBOWA, the cross-lingual loss function is a distance between bag-of-words representations of two aligned sentences.
But as \cite{levy2014neural} showed that the Skip-gram loss function extracts interesting linguistic features, we wanted to use a loss function for the cross-lingual objective that will be closer to Skip-gram than BilBOWA.



Therefore, we introduce a new task, Trans-gram, similar to Skip-gram.
Each English sentence $s_e$ in our aligned corpus $A_{e, f}$ is aligned with a French sentence $s_f$.
In Skip-gram, the context picked for a target word $w_e$ in a sentence $s_e$ is the set of words $c_e$ appearing in the window centered around $w_e$: $s_e[w_e-l:w_e+l]$.
In Trans-gram, the context picked for a target word $w_e$ in a sentence $s_e$ will be all the words $c_f$ appearing in $s_f$.
The loss can thus be written as:
\begin{equation}
    \Omega_{e,f} = \displaystyle 
        \sum_{ (s_e, s_f) \in A_{e, f} }{
            \sum_{w_e \in s_e}{
                \sum_{c_f \in s_f}{
                    - \log \sigma (\vec{w_e} \cdot \cev{c_f})
                }
            }
        }
\end{equation}

This loss isn't symmetric with respect to the languages. We, therefore, use two cross-lingual objectives: $\Omega_{e,f}$ aligning $e$'s target vectors and $f$'s context vectors and $\Omega_{f,e}$ aligning $f$'s target vectors and $e$'s context vectors.
By comparison BilBOWA only aligns $e$'s target vectors and $f$'s target vectors.
The figure \ref{trans_gram} illustrates the four objectives.


Notice that we make the assumption that the meaning of a word is uniformly distributed in the whole sentence.
This assumption, although a naive one, gave us in practice excellent results.
Also our method uses only sentence-aligned corpus and not word-aligned corpus which are rarer.

To add a third language $i$ (e.g. Italian), we just have to add 3 new objectives ($J_i$, $\Omega_{e,i}$ and $\Omega_{i,e}$) to the global loss. If available we could also add $\Omega_{f,i}$ or $\Omega_{i,f}$ but in our case we only used corpora aligned with English.


\section{Implementation}

In our experiments, we used the Europarl \cite{europarlv7} aligned corpora.
Europarl-v7 has two peculiarities: firstly, the corpora are aligned at sentence-level;
secondly each pair of languages contains English as
one of its members: for instance, there is no French/Italian pair.
In other words, English is used as a pivot language.
No bi-lingual lexicons nor other bi-lingual datasets aligned at the
word level were used.

Using only the Europarl-v7 texts as both monolingual and bilingual data, it took 10 minutes to align 2 languages, and two and a half hours to align the 21 languages of the corpus, in a 40 dimensional space on a 6 core computer. We also computed 300 dimensions vectors using the Wikipedia extracts provided by \cite{al2013polyglot} as monolingual data for each language. The training time was 21 hours.


\section{Experiments}

\subsection{Reuters Cross-lingual Document Classification}
We used a subset of the English and German sections of the Reuters RCV1/RCV2 corpora  \cite{reuters2004} (10000 documents each), as in \cite{klementiev2012inducing}, and we replicated the experimental setting.
In the English dataset, there are four topics: CCAT (Corporate/Industrial), ECAT (Economics), GCAT (Government/Social), and MCAT (Markets). We used these topics as our labels and we only selected documents labeled with a single topic.
We trained our classifier on the articles of one language, where each document was represented using an IDF weighted sum of the vectors of its words, we then tested it on the articles of the other language.
The classifier used was an averaged perceptron, and we used the implementation from \cite{klementiev2012inducing}\footnote{Thanks to S. Gouws for providing this implementation}.
The word vectors were computed on the Europarl-v7 parallel corpus with size 40 like other methods. For this task only the target vectors where used.

We report the percentage precision obtained with our method, in comparison with other methods, in Table \ref{results_classification}. The table also include results obtained with 300 dimensions vectors trained by Trans-gram with the Europarl-v7 as parallel corpus and the Wikipedia as mono-lingual corpus.
The previous state of the art results were detained \cite{gouws2014bilbowa} with BilBOWA and \cite{lauly2014autoencoder} with their Bilingual Auto-encoder model.
This model learns word embeddings during a translation task that uses an encoder-decoder approach.
We also report the scores from Klementiev et al. who introduced the task and the BiCVM model scores from \cite{hermann2013multilingual}.

The results show an overall significant improvement over the other methods, with the added advantage of being computationally efficient.

\begin{table*}[t]
\small
\centering
\renewcommand{\arraystretch}{1.3}
\begin{tabular}{l|c|c|c}
\hline
\textbf{Method}  &  \textbf{En $\rightarrow$ De} &  \textbf{De $\rightarrow$ En} &  \textbf{Speed-up in training time} \\
\hline
Klementiev et al. & 77.6\% & 71.1\% & $\times 1$ \\
Bilingual Auto-encoder & \textbf{91.8\%} & 72.8\% & $\times 3$ \\
BiCVM & 83.7\% & 71.4\% & $\times 320$ \\
BilBOWA             & 86,5\% & 75\% & $\times 800$  \\
\hline
\textbf{Trans-gram} & 87,8\% & \textbf{78,7\%} & \textbf{$\times 600$}\\
\textbf{Trans-gram (size 300 vectors EP+WIKI)} & 91,1\% & \textbf{78,4\%}\\
\hline
\end{tabular}
\caption{Comparison of Trans-gram with various methods for Reuters English/German classification}
\label{results_classification}

\begin{tabular}{l||c|c||c|c}
\multicolumn{5}{c}{\vspace{0.1cm}}\\
\hline
\textbf{Method}  &  
\textbf{En $\rightarrow$ Es P@1} &  \textbf{En $\rightarrow$ Es P@5} & 
\textbf{Es $\rightarrow$ En P@1} & \textbf{Es $\rightarrow$ En P@5} \\ 
\hline
Edit distance       & 13\% & 18\% & 24\% & 27\% \\
Bing                & 55\% &      & 71\% &      \\
Translation Matrix  & 33\% & 35\% & 51\% & 52\% \\
BilBOWA             & 39\% & 44\% & \textbf{51\%} & 55\% \\
\hline
\textbf{Trans-gram} & \textbf{45\%} & \textbf{61\%} & 47\% & \textbf{62\%} \\
\hline
\end{tabular}
\caption{Results on the translation task}
\label{results_tr}
\end{table*}

\subsection{P@k Word Translation}
Next we evaluated our method on a word translation task, introduced in \cite{mikolov2013exploiting} and used in \cite{gouws2014bilbowa}. The words were extracted from the publicly available WMT11\footnote{http://www.statmt.org/wmt11/} corpus. The experiments were done for two sets of translation: English to Spanish and Spanish to English.
\cite{mikolov2013exploiting} extracted the top $6K$ most frequent words and translated them with Google Translate.
They used the top $5K$ pairs to train a translation matrix, and evaluated their method on the remaining $1K$.
As our English and Spanish vectors are already aligned we don't need the $5K$ training pairs and use only the $1K$ test pairs.

The reported score, the translation precision $P@k$, is the fraction of test-pairs where the target translation (Google Translate) is one of the $k$ translations proposed by our model. 
For a given English word, $w$, our model takes its target vectors $\vec{w}$ and proposes the $k$ closest Spanish word using the co-similarity of their vectors to $\vec{w}$.
We compare ourselves to the ``translation matrix'' method and to the BilBowa aligned vectors.
We also report the scores obtained by a trivial algorithm that uses edit-distance to determine the closest translation and by the Bing Translator service.


\section{Interesting properties}

\subsection{Cross-lingual disambiguation}

We now present the task of cross-lingual disambiguation as an example of possible uses of aligned multilingual vectors. The goal of this task is  to find a suitable representation of each sense of a given polysemous word. The idea of our method is to look for a language in which the undesired senses are represented by unambiguous words and then to perform some arithmetic operation. 

Let's illustrate the process with a concrete example: consider the French word ``train'', $\wtvTitle{train}{fr}$. The three closest Polish words to $\wtv{train}{fr}$ translate in English into ``now'', ``a train'' and ``when''. This seems a poor matching. In fact, $\wtvTitle{train}{fr}$ is polysemous.
It can name a line of railroad cars, but it is also used to form progressive tenses. The French ``Il est \emph{en train de} manger'' translates into ``he is eat\emph{ing}'', or in Italian ``\emph{sta} mangiando''.

As the Italian word ``sta'' is used to form progressive tenses, it's a good candidate to disambiguate $\wtvTitle{train}{fr}$.
Let's introduce the vector $\vec{v} = \wtv{train}{fr} - \wtv{sta}{it}$.
Now the three polish words closest to $\vec{v}$ translate in English into ``a train'', ``a train'' and ``railroad''.
Therefore $\vec{v}$ is a better representation for the railroad sense of $\wtvTitle{train}{fr}$.

\subsection{Transfer of linguistic features}

Another interesting property of the vectors generated by Trans-gram is the transfer of linguistic features through a pivot language that does not possess these features.


Let's illustrate this by focusing on Latin languages, which possess some features that English does not, like rich conjugations.
For example, in French and Italian the infinitives of ``eat'' are $\wtvTitle{manger}{fr}$ and $\wtvTitle{mangiare}{it}$, and the first plural persons are $\wtvTitle{mangeons}{fr}$ and $\wtvTitle{mangiamo}{it}$. Actually in our models we observe the following alignments:
$\wtv{manger}{fr}   \approx \wtv{mangiare}{it}$ and
$\wtv{mangeons}{fr} \approx \wtv{mangiamo}{it}$.
It is thus remarkable to see that features not present in English match in languages aligned through English as the only pivot language.
We also found similar transfers for the genders of adjectives and are currently studying other similar properties captured by Trans-gram.

\section{Conclusion}
In this paper we provided the following contributions:
Trans-gram, a new method to compute cross-lingual word-embeddings in a single word space;
state of the art results on cross-lingual NLP tasks;
a sketch of a cross-lingual calculus to help disambiguate polysemous words;
the exhibition of linguistic features transfers through a pivot-language not possessing those features.

We are still exploring promising properties of the generated vectors and their applications in other NLP tasks (Sentiment Analysis, NER...).



\bibliographystyle{acl}
\bibliography{main}

\end{document}




\begin{table*}[t]
\centering 
\makebox[\textwidth][c]{
\begin{tabular}{|*{6}{c|}}

    \hline
    \begin{minipage}[t]{0.12 \columnwidth}
        \center\textbf{Method}
    \end{minipage}  &
    \begin{minipage}[t]{0.10 \columnwidth}
        \fontsize{8}{4}\center\textbf{No word alignments required}
    \end{minipage} &
    \begin{minipage}[t]{0.15 \columnwidth}
        \fontsize{8}{4}\center\textbf{No prior on the mapping between target vectors }
    \end{minipage} &
    \begin{minipage}[t]{0.12 \columnwidth}
        \fontsize{8}{4}\center\textbf{No explicit alignments of target vectors}
    \end{minipage} &
    \begin{minipage}[t]{0.15 \columnwidth}
        \fontsize{8}{4}\center\textbf{Computationally efficient}
    \end{minipage} &
    \begin{minipage}[t]{0.12 \columnwidth}
        \fontsize{8}{4}\center\textbf{Can leverage mono-lingual corpus}
    \end{minipage}  \\
    \hline
    Mikolov's Translation Matrix  &  $\times$  &  $\times$  & \checkmark & \checkmark & \checkmark \\
    Faruqui's Translation Matrix   &  $\times$  &  $\times$  & \checkmark & \checkmark & \checkmark \\
    Zou's Translation Matrix       &  $\times$  &  $\times$  & \checkmark & \checkmark & \checkmark \\
    BiCVM         & \checkmark & \checkmark &  $\times$  & \checkmark &  $\times$  \\
    Bilingual Auto-Encoder    & \checkmark & \checkmark &  $\times$  &  $\times$  & $\times$ \\
    Klementiev & \checkmark &  $\times$  & \checkmark &  $\times$  & \checkmark \\
    BilBOWA       & \checkmark & \checkmark &  $\times$  & \checkmark & \checkmark \\
    \hline
    \textbf{Trans-gram}           & \checkmark & \checkmark & \checkmark & \checkmark & \checkmark \\
    \hline
\end{tabular}
}
\caption{Comparative study of each method}
\label{comparative}
\end{table*}

\begin{table*}[t]
\centering
\begin{tabular}{|c|c|}
    \hline
    $\wtvTitle{beau}{fr}$      & $\wtvTitle{belle}{fr}$    \\
    \hline
    $\wtvTitle{bel}{it}$       & $\wtvTitle{bella}{it}$    \\
    $\wtvTitle{lindo}{pt}$     & $\wtvTitle{bonita}{pt}$   \\
    $\wtvTitle{encantador}{es}$& $\wtvTitle{bella}{es}$    \\
    $\wtvTitle{frumos}{ro}$    & $\wtvTitle{adormita}{ro}$ \\
    \hline
\end{tabular}
\caption{
    Closest word in different languages to the given french word.
    To the masculine ``beau'' (pretty) the closest word are also masculine,
    while the closest words to the feminine ``belle'' (pretty) are all feminine.
}
\label{beau}
\end{table*}

\begin{table*}[t]
\centering
\begin{otherlanguage*}{polish}
\begin{tabular}{|*{4}{c|}}
    \hline
    \multicolumn{2}{|c|}{$\wtvTitle{train}{fr}$} & \multicolumn{2}{c|}{$\wtvTitle{train}{fr} - \wtvTitle{sta}{it}$} \\
    \hline
    pl         & translation  & pl      & translation \\
    \hline
    obecnie   & now       & pociąg     & train      \\
    pociąg    & train     & pociągu    & train      \\
    kiedy     & when      & kolejową   & train      \\
    pociągu   & train     & perony     & platforms  \\
    tym       & this      & torów      & tracks     \\
    którym    & where     & wagonów    & wagons bus \\
    na        & on        & tory       & tracks     \\
    aktualnie & currently & torze      & track      \\
    perony    & platforms & trasie     & from       \\
    czym      & what      & prędkością & speed      \\
    \hline
\end{tabular}
\end{otherlanguage*}

\caption{
    Top ten Polish words and their English translations, for the ambiguous French word ``train'' and its disambiguated version.
}
\label{train}
\end{table*}

\begin{table*}[t]
\centering \small
\makebox[\textwidth][c]{
\begin{tabular}{|*{6}{c|}}
    \hline
    fr & it & pt & es & en & de \\
    \hline
    innovation & innovazione & inovação & innovación & non-technological & innovation \\
    innovations & innovativa & criatividade & innovaciones & entrepreneurial & innovationen \\
    créativité & innovazioni & inovações & innovadora & knowledge-driven & unternehmergeist \\
    entrepreneurial & imprenditoriale & inovadora & creatividad & innovations & wissensgesellschaft \\
    innovante & imprenditorialità & tecnologias & tic & information-based & innovations- \\
    entrepreneuriat & pmi & competitividade & tecnologías & innovating & innovationsfähigkeit \\
    compétitivité & innovativo & tic & competitividad & eco-efficiency & unternehmerischer \\
    technologies & creatività & empreendedor & estimulación & research & forschungs- \\
    entreprenariat & imprenditoria & criadora & emprendedor & policy-driven & unternehmerisches \\
    innovant & tecnologie & pci & punteras & internationalisation & wettbewerbsfähigkeit \\
    \hline
\end{tabular}
}
\makebox[\textwidth][c]{
\begin{tabular}{|*{5}{c|}}
    \multicolumn{5}{c}{\vspace{0.1cm}}\\
    \hline
    pl & el & fi & bg & cs \\
    \hline
    innowacji      & \textgreek{καινοτομία}  & kilpailukyvyn & иновациите & inovaci \\
    innowacyjności & \textgreek{καινοτομίας} & teknologian & иновации & inovace \\
    innowacyjność  & \textgreek{καινοτομίες} & kilpailukykyä & иновация & inovací \\
    innowacje      & \textgreek{καινοτομιών} & kilpailukyky & иновацията & inovacím \\
    innowacjom     & \textgreek{γνώσης}      & tutkimukseen & научноизследователската & inovační \\
    innowacyjnej   & \textgreek{επιχειρηματικότητας} & pk-yritysten & предприемаческия & internacionalizaci \\
    innowacjami    & \textgreek{καινοτόμων}  & työllisyyttä & иновационния & znalostní \\
    innowacyjnego  & \textgreek{δημιουργικότητας} & kasvun & иновационен & výzkumu \\
    tworzenia      & \textgreek{καινοτόμος}  & kasvua & развойната & znalostního \\
    kreatywność    & \textgreek{καινοτόμο}   & työpaikkojen & знанието & inovacemi \\
    \hline
    \multicolumn{5}{c}{\vspace{0.1cm}}\\
    \hline
    hu & nl & da & et & ro  \\
    \hline
    innováció & innovatie           & innovation          & innovatsiooni   & inovare \\
    innovációnak & innovaties          & innovationen        & innovatsioonile & inovaţiei \\
    innovációhoz & innovatievermogen   & videnbaseret        & teadmuspõhise   & inovaţie \\
    innovációs & innovatiecapaciteit & nyskabelse          & uuendustele     & inovarea \\
    innovációt & innoverende         & innovationer        & uuendusi        & inovării \\
    kutatás & ondernemersklimaat & innovativ  & innovatsioonipotentsiaali & inovaţia  \\
    tudásba & innovatief & innovations-                 & uuendustest     & cunoaşterii \\
    innováción & innovatiepotentieel & innovationsevne     & uuenduste       & cunoaştere \\
    kutatás-fejlesztés & kennismaatschappij & ivaerksaetterånd     & uuendusteks     & inovări \\
    újításokat & ondernemerschap & innovativt              & innovatsiooniks & inovaţii \\
    \hline
    \multicolumn{5}{c}{\vspace{0.1cm}}\\
    \hline
    sl & sk & lt & sv & lv \\
    \hline
    inovacij & inovácie & naujovių & innovation & jauninājumiem \\
    inovacije & inovácií & inovacijų & nyskapande & inovācijai \\
    inovativnost & inovácii & naujoves & innovationen & inovāciju \\
    inovativnosti & inováciám & naujovėms & innovationer & inovācijas \\
    inovacijam & inováciu & inovacijoms & innovationsförmåga & jauninājumu \\
    inovativnostjo & výskumu & inovacijomis & entreprenörsandan & inovācijās \\
    inovacijske & inovácia & inovacijas & kunskapsbaserad & inovācijām \\
    ustvarjanja & inovačnej & naujovėmis & kunskapsbaserat & jauninājumus \\
    inovacijah & znalostnej & inovacijos & innovationerna & jaunrades \\
    inovacijami & konkurencieschopnosť & naujovės & innovationsområdet & novatorisma \\
    \hline
\end{tabular}
}
\caption{
    For each of the 21 languages of the Europarl, the top ten words closest to $\wtvTitle{innovation}{fr}$
}
\label{innovation}
\end{table*}

\bibliography{main}